\documentclass[10pt,twocolumn,letterpaper]{article}

\usepackage{cvpr}              %
\usepackage[accsupp]{axessibility}
\usepackage[usenames,dvipsnames]{xcolor}
\usepackage{amsmath}
\usepackage{amssymb}
\usepackage{amsfonts}
\usepackage{booktabs}
\usepackage[multidot]{grffile}
\usepackage{array,multirow,graphicx}
\usepackage{multicol}
\usepackage{tikz}
\usepackage{pifont}
\usepackage{subcaption}
\usepackage[pagebackref,breaklinks,colorlinks]{hyperref}

\newcommand{\nada}[1]{}
\def\SelfSR{L1BSR}
\def\DWL{Anchor-Consistency}

\usepackage{tikz}
\usetikzlibrary{spy}

\newcommand{\spiedthree}[2]{\begin{tikzpicture}[every node/.style={inner sep=0,outer sep=0},spy using  outlines={white,magnification=3,size=1.0cm, connect spies}]
\node {\pgfimage[width=\textwidth]{#2}} ;
\spy on #1 in node at (current bounding box.south west) [anchor=south west];
\end{tikzpicture}}

\usepackage[capitalize]{cleveref}
\crefname{section}{Sec.}{Secs.}
\Crefname{section}{Section}{Sections}
\Crefname{table}{Table}{Tables}
\crefname{table}{Tab.}{Tabs.}

\def\cvprPaperID{2} %
\def\confName{EarthVision}
\def\confYear{2023}

\begin{document}

\title{L1BSR: Exploiting Detector Overlap for Self-Supervised Single-Image Super-Resolution of Sentinel-2 L1B Imagery}

\author{Ngoc Long Nguyen$^{1,*}$ \hfill Jérémy Anger$^{1,2,*}$ \hfill Axel Davy$^1$ \hfill  Pablo Arias$^1$ \hfill Gabriele Facciolo$^1$\\
$^1$ Université Paris-Saclay, CNRS, ENS Paris-Saclay, Centre Borelli, France \hspace{0.6cm}$^2$ Kayrros SAS\\
{\small $^*$ These authors contributed equally to this work}.\\
{\tt\normalsize \color{purple} \url{https://centreborelli.github.io/L1BSR/}}\\
}

\twocolumn[{%
\renewcommand\twocolumn[1][]{#1}%
\maketitle
\begin{center}
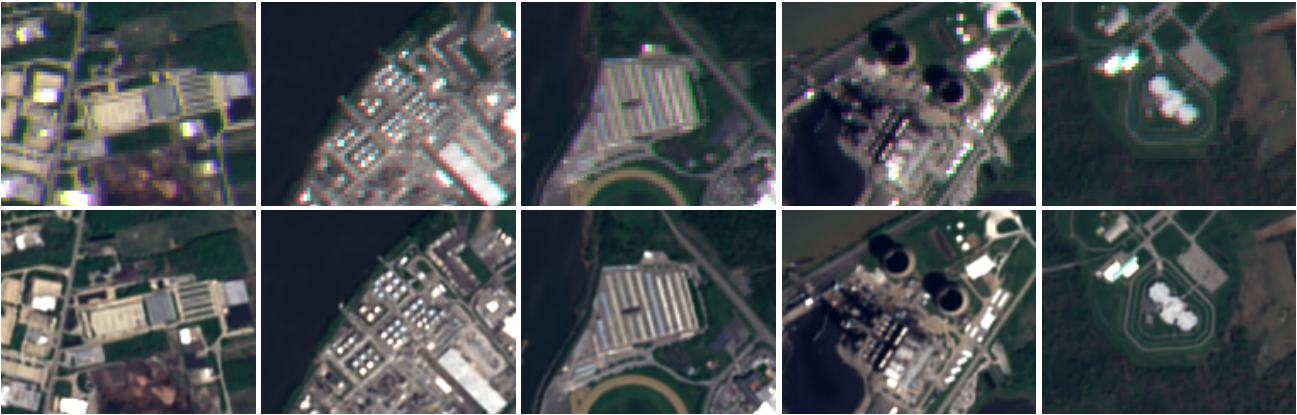

    \def\s{0.193\linewidth}%
    \centering%
    \captionsetup{type=figure}
    \foreach \a in {fig1,fig6,fig7,fig8,fig4}{%
        \includegraphics[width=\s]{./Figures/real_self_sr_results/\a-lr.png}
    }\\[0.1em]%
    \foreach \a in {fig1,fig6,fig7,fig8,fig4}{%
        \includegraphics[width=\s]{./Figures/real_self_sr_results/\a-hr.png}
    }%
    \captionof{figure}{\SelfSR{} produces a 5m high-resolution (HR) output with all bands correctly registered from a single 10m low-resolution (LR) Sentinel-2 L1B image with misaligned bands. Note that our method is trained {on real data} with self-supervision, i.e. without any ground truth HR targets.}
    \label{fig:real_self_sr_results}
\end{center}
}]

\begin{abstract}
High-resolution satellite imagery is a key element for many Earth monitoring applications. Satellites such as Sentinel-2 feature characteristics that are favorable for super-resolution algorithms such as aliasing and band-misalignment. Unfortunately the lack of reliable high-resolution (HR) ground truth limits the application of deep learning methods to this task. In this work we propose L1BSR, a deep learning-based method for single-image super-resolution and band alignment of Sentinel-2 L1B 10m bands. The method is trained with self-supervision directly on real L1B data by leveraging overlapping areas in L1B images produced by adjacent CMOS detectors, thus not requiring HR ground truth. Our self-supervised loss is designed to enforce the super-resolved output image to have all the bands correctly aligned. This is achieved via a novel cross-spectral registration network (CSR) which computes an optical flow between images of different spectral bands. The CSR network is also trained with self-supervision using an Anchor-Consistency loss, which we also introduce in this work. We demonstrate the performance of the proposed approach on synthetic and real L1B data, where we show that it obtains comparable results to supervised methods. 

\end{abstract}

\section{Introduction}
\label{sec:intro}

Earth observation (EO) satellites play a crucial role in our understanding of the Earth systems including climate, natural resources, ecosystems, and natural and human-induced disasters. The Sentinel-2 mission, which is a part of the Copernicus Programme by the European Space Agency (ESA), is considered a significant EO effort alongside other missions such as Landsat. {Sentinel-2} provides optical images of Earth's surface in 13 spectral bands, 4 bands at 10m resolution, 6 bands at 20m, and 3 bands at 60m. The blue (B), green (G), red (R), and near-infrared (N) bands at a ground sample distance (GSD) of 10m/pixel are particularly useful for a variety of applications, including land cover classification, vegetation monitoring, and urban mapping~\cite{drusch2012sentinel}. However, for certain tasks, such as identifying small objects or analyzing fine-scale features, {this} spatial resolution is still inadequate. To address this limitation, super-resolution (SR) techniques can be used to achieve a GSD better than 10m for the RGBN bands. 

SR approaches can be broadly classified into multi-image super-resolution (MISR) and single-image super-resolution (SISR). MISR aims at reconstructing a high-resolution (HR) image from a set of low-resolution (LR) images, typically captured with different viewpoints~\cite{anger2020fast, nguyen2021self, nguyen2022self, lafenetre2023handheld} or at different satellite passes~\cite{martens2019super, arefin2020multi}. If the LR images contain alias and sub-pixel misalignment, they present a perfect scenario for MISR to leverage complementary information in different frames and to recover the true details in the HR output~\cite{nguyen2021self}. %
SISR, on the other hand, is often considered an ill-posed problem due to the potential loss or corruption of high-frequency information caused by factors such as noise, blur, or compression. Nonetheless, a recent study~\cite{nguyen2023role} demonstrates the possibility of SISR for Sentinel-2 10m bands thanks to its unique sensor specifications, namely the inter-band shift and aliasing. The misaligned bands sample the ground at different positions. Since they are correlated each band %
obtains complementary information from the other bands, in a situation similar to a demosaicing problem.

Deep learning (DL) SISR methods currently outperform traditional model-based approaches by a large margin~\cite{wang2020deep}. To date, all learning-based methods for SISR of Sentinel-2 10m bands have used supervised training, which penalizes a loss between the HR image predicted by the network and a ground truth HR image. Some studies attempt to train a SISR model on a simulated dataset where LR images are generated using a pre-defined degradation model~\cite{pineda2020generative}. However, the performance may drop substantially if the real low-resolution input deviates from the {simulated} degradation model. Other {works} use real HR {images acquired by other} satellites to directly supervise the SR of Sentinel-2~\cite{galar2020super,nguyen2023role}. However, obtaining the HR ground truth images can be costly. In addition, the use of HR images from different satellites {introduces} challenges such as {spectral response discrepancies, and acquisition viewpoint and time differences}, which complicate the process of dataset creation and {negatively impact performance}.

\begin{figure}
    \centering
    \includegraphics[width=0.97\linewidth]{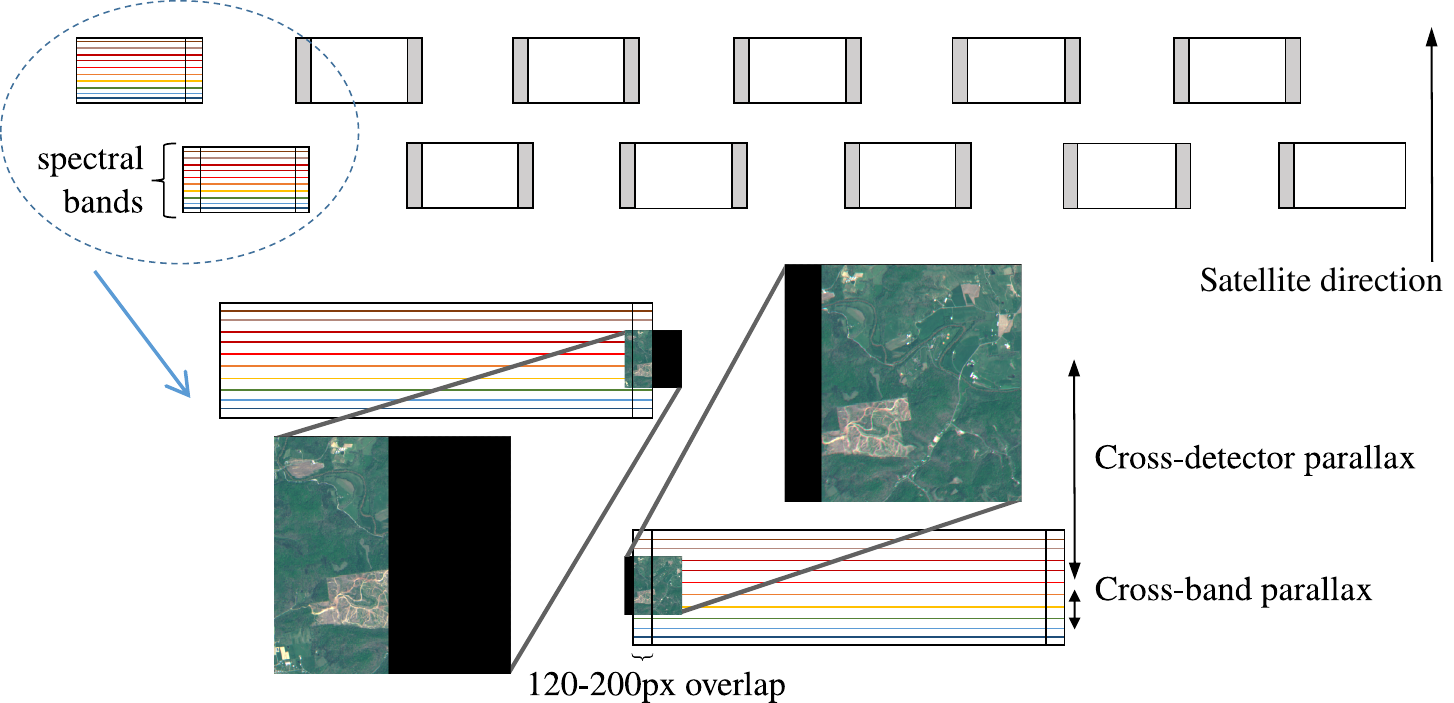}
    \caption{Sensor layout of the Sentinel-2 MSI (figure adapted from~\cite{gascon2017copernicus}). {The push-broom acquisition is done in the vertical direction.} The overlap between detectors provides two near-simultaneous observations of the scene.}
    \label{fig:s2-sensor-layout}
\end{figure}

A promising direction is to use self-supervised learning techniques, which have been applied to multi-image restoration tasks such as video/burst denoising and demosaicing~\cite{ehret2019model,ehret2019join,dewil2021self,yu2020joint,sheth2021unsupervised}, and recently to MISR in the context of push-frame satellites~\cite{nguyen2021self, nguyen2022self} that acquire bursts of images at high frame rate.
These techniques exploit redundant information from multiple observations: one of the degraded frames in the input sequence is withheld from the network and used as label instead of the ground truth. As a consequence they require at least two degraded observations from the same HR signal.

In this work we leverage a unique feature of the Sentinel-2 hardware design that enables self-supervised training of single-image super-resolution (Fig.\ref{fig:real_self_sr_results}). %
Sentinel-2 is equipped with a MultiSpectral Instrument (MSI) that has 12 detectors capturing information in the in the visible and near infrared (VNIR) wavelength range.%
These detectors operate in a push-broom fashion, scanning the image line-by-line as the satellite moves over the ground (as illustrated in Fig.\ref{fig:s2-sensor-layout}). Of note, 
{adjacent} 
detectors share a 2 km overlap area across the track (120 to 200 pixels), which offers opportunities for self-supervised image restoration techniques. 

These overlapping regions are only available in early products in the Sentinel-2 processing pipeline, such as the level-1B (L1B) products, which present a significant inter-band parallax due to the hardware design of the detectors. This parallax, while beneficial for the super-resolution task, is undesirable for human interpreters, which is why it is removed later in the pipeline (e.g. L1C products) using camera calibration information to align the different bands.

\paragraph{Contributions.}
In this work we propose \SelfSR{}, a novel 
self-supervised method for SISR and band alignment of Sentinel-2 L1B RGBN bands. 
The method is trained directly on real L1B data 
using image crops contained in the {detector} overlap regions.
One of the overlapping L1B crops is given as input to our network, and the other is used as target in the loss. The network is tasked to generate a super-resolved image such that, when properly aligned and downsampled, matches the target {10m} LR crop~\cite{nguyen2021self,nguyen2022self}. 
It should be noted that once trained, as a SISR method, our network has the capability to produce {5m} HR images throughout the image domain, rather than just at the overlapping regions.

Our self-supervised loss is designed to enforce the super-resolution network to output an HR image with the bands correctly aligned. This is achieved by aligning all bands of the target image with the green channel of the super-resolved output.

To that aim, as a second contribution of our work, we present a novel cross-spectral registration (CSR) method which allows to compute an optical flow between images of different spectral bands. 
To train the CSR network we propose \DWL{}, a simple yet effective self-supervised loss for cross-spectral registration.
Our self-supervised cross-spectral registration simultaneously learns to handle all possible band combinations.
We use our CSR network only during training of the SR network as part of the self-supervised loss. 
Once the reconstruction network is trained, the cross-spectral registration network is no longer required. 
The reconstruction network can directly generate high-quality HR images with correctly registered bands (as shown in Fig.\ref{fig:real_self_sr_results}) without requiring an explicit alignment step, nor calibration information. 

We validate our contributions with an empirical study on a synthetic dataset obtained from L1C products~(Sec.~\ref{sec:experiments}), designed to model the main characteristics of Sentinel-2 {L1B} data. We show that our \SelfSR{} network as well as our cross-spectral registration module trained with the proposed self-supervision strategy attain performance on par with those obtained with supervised training.

We train our self-supervised method on a dataset of 3740 pairs of L1B RGBN overlapping {crops} (Sec.~\ref{sec:l1bdataset}) and compare with a supervised method designed for Sentinel-2 L1C~\cite{nguyen2023role} in Sec.~\ref{sec:qualitative}. %
It is worth noting that our training dataset, which can find applications in various image restoration and cross-spectral registration tasks, will be soon available on our project website.%

\section{Related Work}
\label{sec:related}

\paragraph{SISR for Sentinel-2}
Early research on SISR of Sentinel-2 images primarily focused on pan-sharpening the lower-resolution (20m and 60m) bands to create a uniform 10m GSD data cube~\cite{lanaras2018super, gargiulo2019fast}. Recent years have seen an increased interest in enhancing the resolution of the Sentinel-2 10m bands. Some studies ~\cite{pineda2020generative} generated synthetic LR-HR pairs to train SISR models, but these models tend to suffer from generalization issues~\cite{cai2019toward}. Another trending approach is to directly supervise the SISR of Sentinel-2 using another high-resolution satellite such as PlanetScope~\cite{nguyen2023role, galar2020super, zabalza2022super}, VEN\textmu S~\cite{michel2022sen2venmus}, or WorldView~\cite{salgueiro2020super}. However, creating the training dataset for these approaches requires significant engineering work due to the radiometric and geometric differences between the two constellations.

\paragraph{Self-supervised SR} Self-supervised and unsupervised learning are promising approaches to avoid the need for large labeled datasets in SR tasks. ZSSR~\cite{shocher2018zero} and MZSR~\cite{soh2020meta} have been proposed to model image-specific LR-HR relations during the testing phase using example pairs generated from the LR test image and its degraded version. Although the idea is interesting, it may not be practical to train on each test image.
Another approach is to use cycle-consistency and adversarial losses~\cite{yuan2018unsupervised,kim2020unsupervised,lugmayr2019unsupervised, wang2021unsupervised} to train a neural network without requiring pairs of LR-HR images. %
However, these GAN-based models are prone to producing hallucinations, which may not be acceptable for certain applications.

Our SISR method is both fully self-supervised and free from hallucinations. We drew inspiration from the Noise2Noise framework~\cite{lehtinen2018noise2noise}, which introduced a pioneering approach to train a neural network for image denoising task using pairs of noisy images instead of pairs of noisy-clean images. The key idea behind Noise2Noise is that when comparing pairs of noisy images with independent noise realizations, the network learns to identify the underlying patterns in the noise and removes them accordingly. Similarly by comparing pairs of overlapping L1B images, our network can learn to recognize the aliasing patterns in each LR image and leverage them to recover the high-frequency details in the HR. The closest work to ours is the DSA method~\cite{nguyen2021self}, which addresses MISR for SkySat imagery. During training, DSA hides the LR reference image and asks the network to produce a HR image from the other $n-1$ images such that after downsampling, it coincides with the reference image. Our work can be seen as the SISR version of DSA. However, our network also learns to perform implicit cross-spectral registration at inference time, which is a challenging and compelling task by itself. 

\paragraph{Cross-spectral registration}
Cross-spectral registration refers to the process of aligning two or more images that are captured using different sensors or imaging modalities. Cross-spectral registration has become increasingly important in various fields such as remote sensing~\cite{ye2017robust, pielawski2020comir}, medical imaging~\cite{lee2009learning}, and computer vision~\cite{arar2020unsupervised} as it allows for the integration of information from different spectral bands, thereby yielding richer scene representations. While increasing efforts have been made in the past few years to improve the performance of cross-spectral registration, this still remains an open problem~\cite{jiang2021review}. Feature-based methods~\cite{ye2017robust, liu2018novel} involve identifying distinctive features, such as edges or corners, in both images and then matching them to establish correspondences. Intensity-based methods~\cite{cao2020boosting, zhang2018mutual} rely on the similarity of the pixel values in both images. Examples of intensity-based methods include normalized cross-correlation, mutual information, and phase correlation. In recent years, deep learning-based methods have also been explored and achieved state-of-the-art (SOTA) performance. Most DL studies including ~\cite{pielawski2020comir, arar2020unsupervised, walters2021there, xu2022rfnet} have employed image-to-image translation~\cite{isola2017image} techniques to map two images to the same image space and then register them accordingly. However, these methods require extensive work in designing models and sophisticated training losses. In contrast, we propose \DWL{} a novel and straightforward loss for training a cross-modal registration network. Notwithstanding its simplicity, our method provides a strong baseline for the task. In addition, our loss can also be easily integrated into existing frameworks to improve consistency or used as a quantitative metric for evaluating cross-modal registration quality.

\section{Proposed Method}
\label{sec:method}
Our primary aim is to leverage detector overlaps in Sentinel-2 L1B images to learn to recover high-frequency details hidden in its misaligned bands.
{Note that the maximum attainable resolution is capped by the spectral decay of the blur kernel resulting from the sensor's pixel integration and the camera optics, which imposes a frequency cutoff beyond which there is no usable high frequency information~\cite{baker02limits}. For this reason, our aim in this work is to increase the resolution by a factor 2.}
In this section, we first present an overview of our proposed \SelfSR{} framework (Sec.~\ref{sec:archi}). Then, we describe
our self-supervised losses in Sec.~\ref{sec:selflearing} and provide details about the training %
in Sec.~\ref{sec:training}.

 Throughout the text, we denote by $I_t, t\in\{0,1\}$ the two 4-channel %
 overlapping images. We refer to $I_0$ as the input (or reference image) for the SR task and $I_1$ as the target for our self-supervised losses, which are explained in more detail in Sec.~\ref{sec:selflearing}. $I_{t,i}$ is the grayscale image extracted from the channel $i$ of $I_t$, where $i\in \{b,g,r,n\}$ and $b$, $g$, $r$, and $n$ stand for the blue, green, red, and near-infrared channels, respectively.

\subsection{Architecture}
\label{sec:archi}
\begin{figure}
    \centering
    \includegraphics[width=0.98\linewidth]{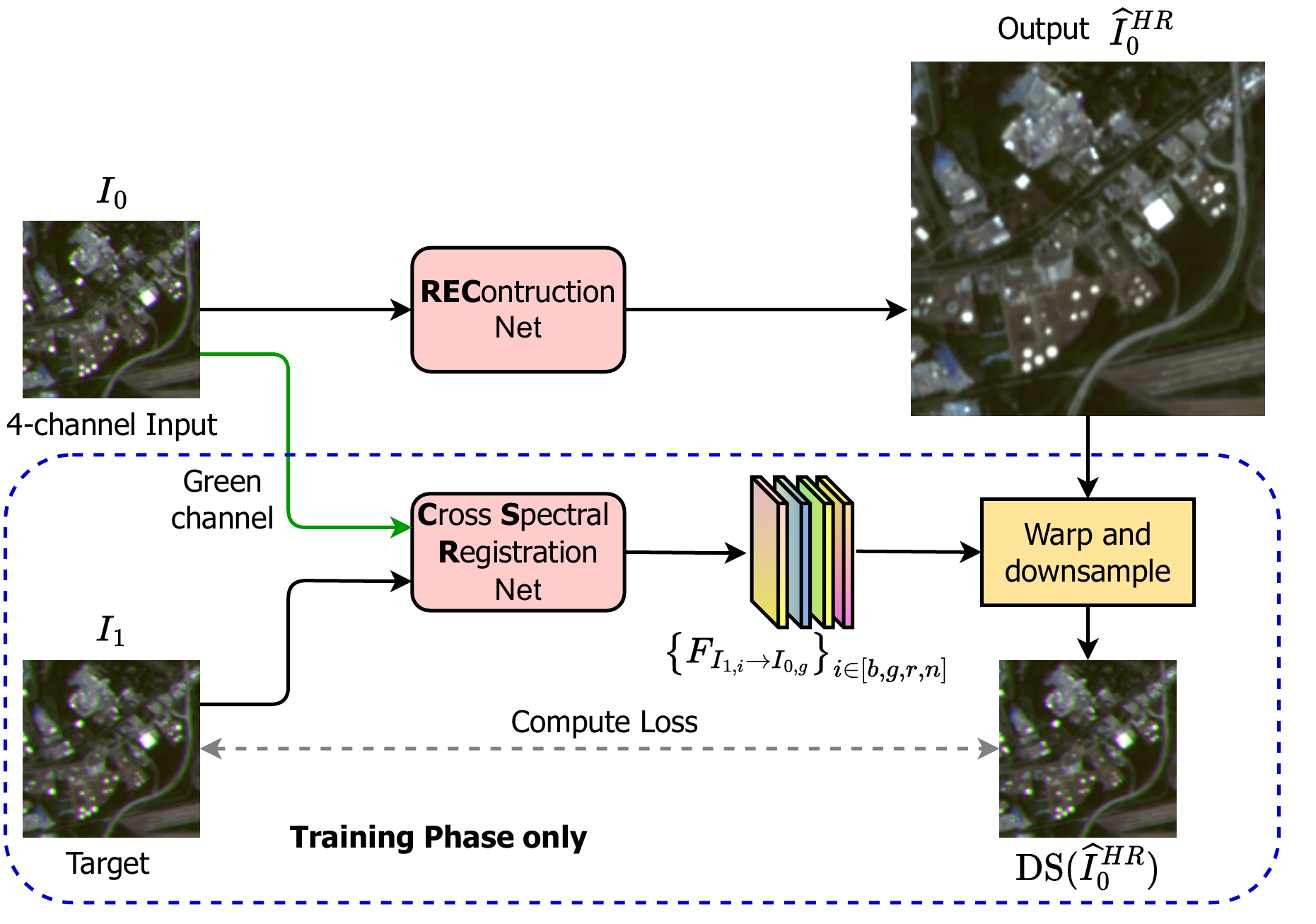}
    \caption{Overview of our proposed self-supervised \SelfSR{} framework for Sentinel-2 L1B at training time. The depicted loss represents the self-supervision  term $\ell_{\text{Self-SR}}$~\eqref{eq:ssloss}. Note that at inference time, only one input and the reconstruction module are required.} 
    \label{fig:Architecture}
\end{figure}

Our proposed \SelfSR{} framework (Fig.~\ref{fig:Architecture}) consists of two main components: a cross-spectral registration network (\textbf{CSR}) and a reconstruction network (\textbf{REC}). The \textbf{CSR} module {computes}
dense correspondences between $I_{0,g}$ and all the bands of $I_1$. By utilizing these motion fields during training, the \textbf{REC} network learns to produce a HR output $\widehat I_0^{HR}$ with all four channels aligned with $I_{0,g}$. Of note, \underline{the \textbf{CSR} module is not used at test time}. Instead, the \textbf{REC} network performs cross-channel registration implicitly.

\paragraph{Reconstruction Network}
Our \textbf{REC} network is built on the Residual Channel Attention Networks (RCAN) architecture~\cite{zhang2018image}, which has been shown to achieve state-of-the-art performance on many image restoration tasks.
We chose this architecture mainly due to its channel attention mechanism, which can be viewed as a weighting function that enables \textbf{REC} to selectively focus on informative channels in the feature space and disregard irrelevant ones. %

The reconstruction network takes a LR image $I_0\in\mathbb R^{H\times W\times 4}$ with misaligned bands as input and produces a super-resolved output by a factor of two $\widehat I_0^{HR}$ with all four bands aligned with the green channel of the input image
\begin{equation}
    \widehat I_0^{HR} = \textbf{REC}(I_0; \Theta_{\textbf{REC}}) \in \mathbb{R}^{2H\times 2W\times 4},
\end{equation} 
where $\Theta_{\textbf{REC}}$ denotes the network parameters. We opt for the default RCAN configuration to strike a balance between computational efficiency and performance. Overall, the \textbf{REC} network contains 10 residual groups, each with 20 residual channel attention blocks (RCAB), and a global skip connection. Each RCAB is a combination of a residual block and a channel attention layer implemented using a ``squeeze-and-excitation" technique~\cite{hu2018squeeze}. %
The number of feature channels is fixed to 64 across all layers. %

It is important to highlight that the task our \textbf{REC} network {must} accomplish is particularly challenging, as it involves both super-resolution and cross-spectral registration at the same time. To tackle this problem, we incorporate a dedicated \textbf{CSR} module into the training process, which enables the \textbf{REC} network to learn efficiently the task in a self-supervised way. %

\paragraph{Cross-Spectral Registration Network}
The \textbf{CSR} module is instrumental during training. We use it to train the \textbf{REC} network to produce an HR output where all bands are aligned to the green {one} (as justified in Sec.~\ref{subsec:csr}). By having to align the channels, it becomes easier for the \textbf{REC} network to learn inter-band correlations, %
and thereby to leverage the complementary information in each band. %

The cross-spectral registration network takes %
{any two} spectral bands of Sentinel-2 L1B images $I_{\cdot,i}$ and $I_{\cdot,j}$, with $i, j \in \{b,r,g,n\}$ as input and produces a dense correspondence between them
\begin{equation}
    F_{I_{\cdot,j} \to I_{\cdot,i}} \!= \textbf{CSR}(\bar I_{\cdot,i}, \bar I_{\cdot,j}; \Theta_{\textbf{CSR}}) \in [-R,R]^{H \times W \times 2},
\end{equation} 
where $\Theta_{\textbf{CSR}}$ denotes the parameters of $\textbf{CSR}$, {and $\bar I$ is the normalization of image $I$ according to its mean and standard deviation}. The network is trained with a maximum motion range of $[-R,R]^2$ (with $R = 10$ pixels). Note that $I_{\cdot,i}$ and $I_{\cdot,j}$ should represent the same scene and be extracted either from the same image or from two overlapping images. The \textbf{CSR} follows a simple U-Net architecture~\cite{ronneberger2015u} with 4 scales to increase the receptive field of the convolutions. %

\begin{figure}
    \centering
    \includegraphics[width=0.98\linewidth]{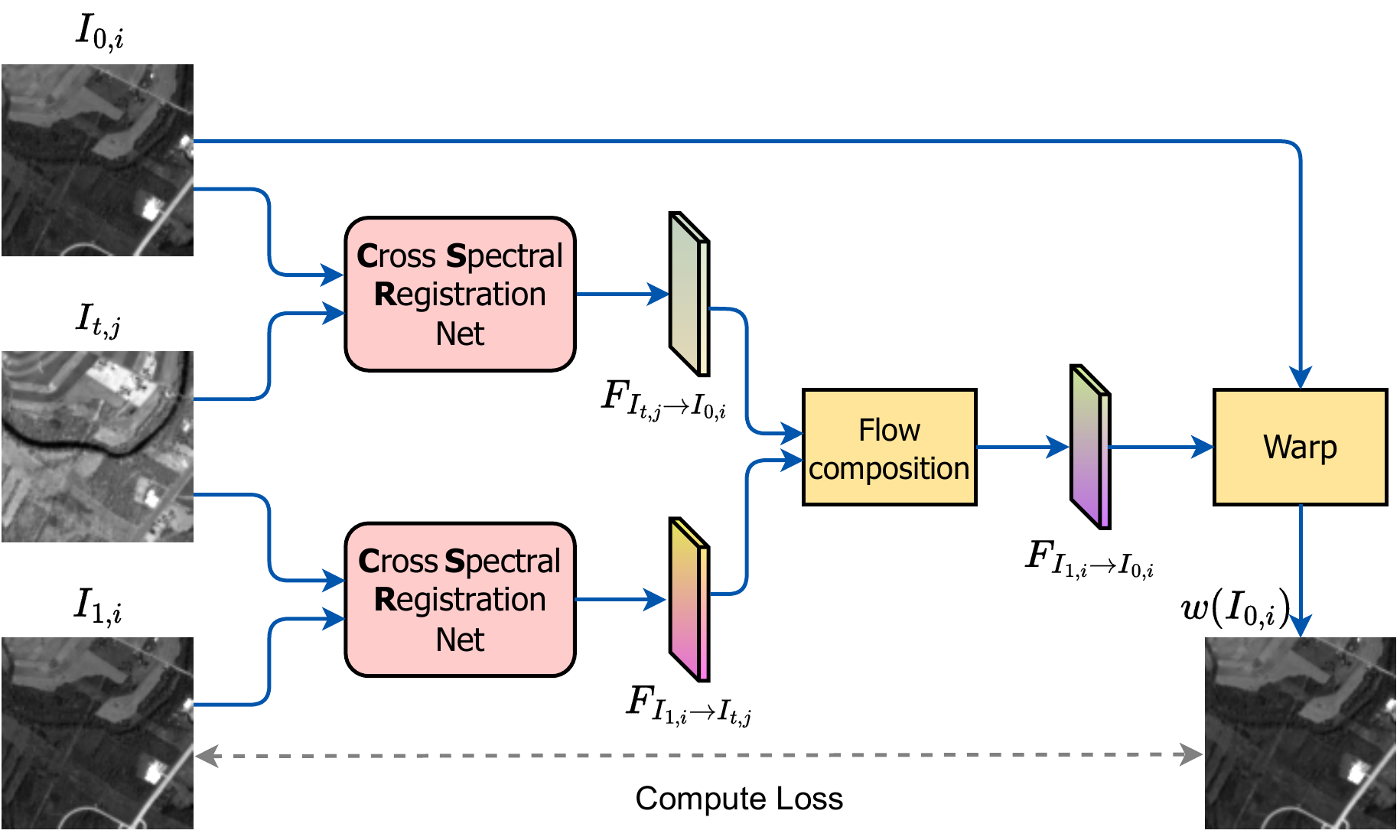}
    \caption{Training setup of our proposed cross-spectral registration (\textbf{CSR}) module. The motion from $I_{0,i}$ to $I_{1,i}$ via the anchor $I_{t,j}$ should represent the direct motion between them. The depicted loss represents the \DWL{} loss~\eqref{eq:DWL}.}
    \label{fig:DW}
\end{figure}

\subsection{Self-supervised learning}
\label{sec:selflearing}
Our framework is trained in a fully self-supervised manner, i.e. without requiring ground truth. In this section, we describe our self-supervised losses that are used to train the \textbf{REC} and the \textbf{CSR} modules.
\paragraph{Self-SR Loss} 
By utilizing the motion fields between the green %
{band} of $I_0$ and all %
{bands} of $I_1$ computed by \textbf{CSR}, \textbf{REC} can produce a HR output $\widehat I_0^{HR}$ with all bands correctly registered to $I_{0,g}$. We achieve this by minimizing the Self-SR loss:
\begin{equation}
    \ell_{\text{Self-SR}} = \|\Pi_2 \omega_{2}(\widehat I_0^{HR}, F_{I_1 \to I_{0,g}})- I_1\|_1,
    \label{eq:ssloss}
\end{equation}
where $\Pi_{2}$ denotes the subsampling operator and $\omega_{2}(-, F)$ computes a bicubic sampling ({Pullback}) in the HR domain using the LR flow $F$. $F$ denotes actually 4 optical flows, one for each band in $I_1$.
{The} operator $\omega_2$ is equivalent to a backward warping with an upscaled version of the flow $2F$.

The Self-SR loss forces $\omega_{2}(\widehat I_0^{HR}, F_{I_1 \to I_{0,g}})$ to {be aligned} to $I_1$, resulting in the requirement that all bands of the output $\widehat{I}_0^{HR}$ be registered with $I_{0,g}$. Following the work of~\cite{nguyen2021self}, we can also incorporate a blur kernel $k$ into the Self-SR loss to directly produce a sharp HR image
\begin{equation}
    \ell_{\text{Self-SR}}^* = \|{\Pi_2}\omega_{2}(\widehat I_0^{HR} \ast k, F_{I_1 \to I_{0,g}})- I_1\|_1.
    \label{eq:ssloss_deconv}
\end{equation}
During training, we randomly choose the reference and the target between the two overlapping images. At inference time, we only need one single LR input and the \textbf{REC} network to obtain a high-quality HR output.

\paragraph{\DWL{} Loss}
The \textbf{CSR} network is trained with self-supervision. 
Fig.~\ref{fig:DW} illustrates our training setup for the \textbf{CSR}. For that we need 3 images $I_{0,i}$, $I_{t,j}$, and $I_{1,i}$. The image $I_{t,j}$ serves as an anchor image extracted from either the $I_0$ or $I_1$ (i.e. $t \in \{0,1\}$) but may come from different spectral band than the two other images (i.e. $j \neq i$). We compute the motion fields between these 3 images in two steps:
\begin{equation}
\begin{aligned}
    F_{I_{t,j} \to I_{0,i}} &= \textbf{CSR}(\bar I_{0,i}, \bar I_{t,j}), \\
    F_{I_{1,i} \to I_{t,j}} &= \textbf{CSR}(\bar I_{t,j}, \bar I_{1,i}).
    \label{eq:CSR_setup}
\end{aligned}
\end{equation}

These two motion fields should be consistent in such a way that their composition enables alignment between $I_{0,i}$ and $I_{1,i}$:
\begin{equation}
    \widehat F_{I_{1,i} \to I_{0,i}} = F_{I_{1,i} \to I_{t,j}} \circ F_{I_{t,j} \to I_{0,i}}.
    \label{eq:flow_composition}
\end{equation} 
The \DWL{} loss constrains that the motion from $I_{0,i}$ to $I_{1,i}$ via the anchor should represent the direct motion between them. %
\begin{equation}
    \ell_{\text{\DWL}} = \|\omega(I_{0,i}, \widehat F_{I_{1,i} \to I_{0,i}}) - I_{1,i}\|_1,
    \label{eq:DWL}
\end{equation} 
where %
$\omega$ is the classic pullback operator. %

During training, we randomly choose the reference and the target between the two overlapping images. The spectra $i$ and $j$ are also picked arbitrarily in $\{b, r, g, n\}$. It is important to note that the case where $i$ and $j$ are identical is also {considered} for \textbf{CSR} to learn to register images %
{of the same band}.

\subsection{Training details}
\label{sec:training}
We %
{train first} the \textbf{CSR} network, %
{as it is important to ensure} that the \textbf{CSR} output %
can be effectively utilized by the \textbf{REC} network. %
We %
{employ} the \DWL{} loss~\eqref{eq:DWL} to train the network, with weights initialized using Xavier's initialization~\cite{glorot2010understanding}. %
We set the batch size to $64$ and used Adam~\cite{kingma2014adam} with the default PyTorch parameters and a learning rate of $5e-5$ to optimize the loss. The training converged after 200k iterations and took approximately 24 hours on a single NVIDIA V100 GPU.

The second phase consists of training the \textbf{REC} network using the Self-SR loss~\eqref{eq:ssloss} and the trained \textbf{CSR}. We train \textbf{REC} on LR crops of size $96\times96\times4$ pixels and validate on LR images of size $256\times256\times4$ pixels. We set the batch size to 16 and optimize the loss using the Adam optimizer with default parameters. Our learning rate is initialized to $5e-5$ and decayed by a factor of 0.6 every 12k iterations. The training converges after 60k iterations and takes about 20 hours to complete on a single NVIDIA V100 GPU. We apply data augmentation (DA) techniques such as flips and rotations. %
{The \textbf{CSR} network is %
{fixed} during the training of the \textbf{REC} network.}

\section{Experiments}
\label{sec:experiments}
In this section, we present experimental results that demonstrate the effectiveness of our fully self-supervised approach for {Sentinel-2 SISR}. {To this aim}, we conduct experiments on both real Sentinel-2 L1B data and a simulated dataset that we generated from Sentinel-2 L1C products. Through extensive ablation studies and quantitative analyses, we aim to {demonstrate} the efficacy of our method in addressing the challenges posed by the cross-spectral registration and SISR tasks in Sentinel-2 imagery. Additionally, we compare our self-supervised approach to {a} state-of-the-art supervised SISR method.

\subsection{Simulated dataset}
\label{sec:simulated-ds}

The simulated dataset used in our experiments was generated from 20 Sentinel-2 L1C products, with 18 used for training and 2 for testing. 
The products were extracted from 5 different continents in both summer and winter seasons to ensure geographic and radiometric diversity. For the training set, we selected 6,998 crops, each with a size of $512\times512\times4$ pixels. For the testing set, we selected 184 crops of the same size.
{Provided that Sentinel-2 imagery contains significant alias~\cite{nguyen2023role} which is unsuitable to use as ground truth HR,} we first applied a Gaussian kernel with $\sigma=0.7$ to each crop to remove some aliasing, %
{approximating the effect of an} optical blur. The ground truth images $I_0^{HR}$ and $I_1^{HR}$ were then generated by applying 2 random homography transformations $\mathcal{H}_0$ and $\mathcal{H}_1$ to the blurred HR image (denoted as $B^{HR}$). These ground truth HR should be aligned to $I_{0,g}$ and $I_{1,g}$, respectively. We also simulated band-misalignment in the LR by applying a small homography transformation, where the translation component is dominant. Additionally, a little Gaussian noise (0.1\%) was added to the LR to match the Sentinel-2 noise level. Overall, the simulation process can be summarized as follows:
\begin{equation}
\label{eq:simulated}
\begin{aligned}
    I_t^{HR} &= \mathcal{H}_t (B^{HR}),  \qquad t \in \{0,1\} \\
    I_{t,g} &= \Pi_{2}(I_{t,g}^{HR}) + n_g  , \\
    I_{t,i} &= \Pi_{2}\left((\mathcal{H}_{t,i}\circ \mathcal{H}_t) (B_i^{HR})\right) + n_i  , \quad i \neq g,
\end{aligned}
\end{equation}
where $\mathcal{H}_{t,i}$ ($i\in \{b,r,n\}$) is a translation-dominant homography modeling the band-misalignment between $I_{t,g}$ and $I_{t,i}$. $n_i$ models the noise in the Sentinel-2 L1B. The largest distortion between 2 bands of 2 images can be up to 10 pixels. To enable diverse ablation studies for both the \textbf{CSR} and the \textbf{REC} networks, the homographies are stored as ground truth flows.

The Sentinel-2 L1C products are derived from the L1B products by the Ground Segment. During this process, the bands are aligned and resampled using camera altitude and geometric models. However, due to imperfect parameter estimation, there may be residual shifts between the bands. %
These shifts are typically less than 0.3 pixels~\cite{gascon2017copernicus}.

\subsection{Cross-spectral registration}
\label{subsec:csr}
Table~\ref{tab:csr} reports the mean absolute pixel error of the self-supervised and supervised \textbf{CSR} networks on our test set when aligning the reference bands $I_{0,i}$ to the target bands $I_{1,j}$ with $i, j \in \{b,g,r,n\}$. The first half of the table shows the performance of the self-supervised \textbf{CSR} network, where the diagonal entries correspond to the same-band registration. The self-supervised \textbf{CSR} network performs exceptionally well for the RGB bands, in particular for the green band, exhibiting low mis-registration (less than 0.04 pixel), indicating a high correlation between the RGB bands. However, the registration between NIR and RGB is much more challenging, with an error around 0.1 pixel, which is twice as large as that of the RGB bands, suggesting a much lower correlation between NIR and RGB bands.

To validate the effectiveness of our self-supervised approach, we also performed supervised training of \textbf{CSR} on the same training set, where we penalized the error between the output flows and the ground truth flows obtained from the homographies. The second half of Table~\ref{tab:csr} presents the results of the supervised model over the test set. The table shows a small gap between the performance of the two models, with a maximum mean error of 0.03 pixels for RGB and 0.09 pixels for NIR in the supervised setting, compared to 0.04 pixels and 0.11 pixels for self-supervision. Overall, the self-supervised \textbf{CSR} approach performs well for many applications, without requiring any ground truth flows, knowledge of the optical instrument or scene modeling.

\begin{table}[t!]
    \centering
    \caption{Cross-spectral registration error (in pixel) of our self-supervised and supervised \textbf{CSR} networks over the synthetic test set ($184 \times 256\times 256\times 4$ pixels). The score of same-band registration is highlighted in {\bf bold}.}
    \begin{tabular}{llcccc}
\toprule
                && \multicolumn{4}{c}{Target bands}\\ 
                \cmidrule{3-6}
&Ref. bands      & B     & G     & R     & N \\
\midrule
\parbox[t]{1mm}{\multirow{4}{*}{\rotatebox[origin=c]{90}{\scriptsize Self-supervised}}} 
&B               & \textbf{0.026}& 0.035 & 0.039 & 0.106   \\
&G               & 0.034 & \textbf{0.026}& 0.038 & 0.092   \\
&R               & 0.035 & 0.037 & \textbf{0.026}& 0.104   \\
&N               & 0.100 & 0.086 & 0.098 & \textbf{0.027}  \\
\midrule
\parbox[t]{1mm}{\multirow{4}{*}{\rotatebox[origin=c]{90}{\scriptsize Supervised}}} 
&B               & \textbf{0.016}& 0.028 & 0.029 & 0.088   \\
&G               & 0.027 & \textbf{0.014}& 0.027 & 0.076   \\
&R               & 0.029 & 0.027 & \textbf{0.017}& 0.090   \\
&N               & 0.083 & 0.072 & 0.081 & \textbf{0.017}  \\
 \bottomrule
    \end{tabular}
    \label{tab:csr}
\end{table}
\subsection{Multi-band super-resolution}

We conducted ablation experiments using the proposed synthetic L1C dataset to evaluate the performance of our self-supervised reconstruction network. 
{We found that the residual misalignment in the L1C product affects PSNR measurements: a super-resolved result with well-aligned bands will be slightly misaligned with respect to the ground truth. To avoid this, we align each band of the ground truth to the corresponding band of the super-resolved output before computing the PSNR (note that this alignment is between images of the same band). We use a classical TV-L1 optical flow~\cite{perez2013tv}, setting the weight of the data attachment term to 0.3. 
The PSNRs shown in Tables \ref{tab:SR_ablation_G} and \ref{tab:SR_ablation_G_alignment} were computed in this way.}

First, we studied the impact of the number of input bands on the reconstruction quality. Table~\ref{tab:SR_ablation_G} shows the PSNR results for four different networks trained with different input bands. 
We observed a significant improvement in performance as we increased the number of bands provided to the \textbf{REC} network, consistent with previous findings~\cite{nguyen2023role}. 

\begin{figure*}[t]
    \centering
    \def\s{0.162}
    \def\pos{(-0.73,0.65)} %
    \def\posi{(0.4,0.93)} 
    \begin{subfigure}{\s\linewidth}%
        \spiedthree{\pos}{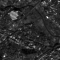}%
        \caption*{LR (Green)}%
    \end{subfigure}
    \begin{subfigure}{\s\linewidth}%
        \spiedthree{\pos}{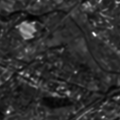}%
        \caption*{G 42.90dB}%
    \end{subfigure}
    \begin{subfigure}{\s\linewidth}%
        \spiedthree{\pos}{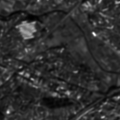}%
        \caption*{BG 46.71dB}%
    \end{subfigure}
    \begin{subfigure}{\s\linewidth}%
        \spiedthree{\pos}{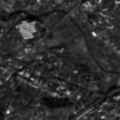}%
        \caption*{BGR 48.24dB}%
    \end{subfigure}
    \begin{subfigure}{\s\linewidth}%
        \spiedthree{\pos}{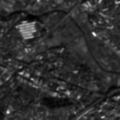}%
        \caption*{BGRN 49.42dB}%
    \end{subfigure}
    \begin{subfigure}{\s\linewidth}%
        \spiedthree{\pos}{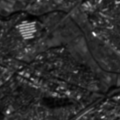}%
        \caption*{HR (Green)}%
    \end{subfigure}

    \begin{subfigure}{\s\linewidth}%
        \spiedthree{\posi}{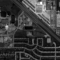}%
        \caption*{LR (Green)}%
    \end{subfigure}
    \begin{subfigure}{\s\linewidth}%
        \spiedthree{\posi}{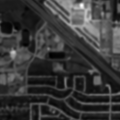}%
        \caption*{G 40.62dB}%
    \end{subfigure}
    \begin{subfigure}{\s\linewidth}%
        \spiedthree{\posi}{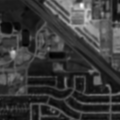}%
        \caption*{BG 44.20dB}%
    \end{subfigure}
    \begin{subfigure}{\s\linewidth}%
        \spiedthree{\posi}{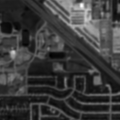}%
        \caption*{BGR 45.85dB}%
    \end{subfigure}
    \begin{subfigure}{\s\linewidth}%
        \spiedthree{\posi}{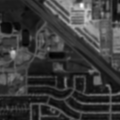}%
        \caption*{BGRN 46.77dB}%
    \end{subfigure}
    \begin{subfigure}{\s\linewidth}%
        \spiedthree{\posi}{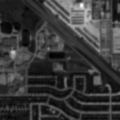}%
        \caption*{HR (Green)}%
    \end{subfigure}
    
    \vspace{-.5em}
    
    \caption{SR reconstruction of the green band when trained jointly with different input spectral bands. We observe that the other bands (B, R, N) provide valuable information for improving the reconstruction of the green band.}
    \label{fig:MCSR}
\end{figure*}

Furthermore, Fig.~\ref{fig:MCSR} illustrates the performance improvements in restoring the G band by providing different bands as input. As more bands are used, the network is able to restore aliased patterns into the true pattern and reach higher signal-to-noise ratio.

Secondly, we compared the performance of our self-supervised framework against a supervised training approach using the synthetic L1C dataset. The supervised training minimized an $L_1$ loss between the restoration and the ground truth HR image.
Table~\ref{tab:SR_ablation_G_alignment} shows the per-band mean PSNRs over the test set. {We observe a significant PSNR gap in favour of the proposed self-supervised \SelfSR{} method ($0.7$dB for the visible bands and $0.3$ for the near-infrared). This is rather unexpected. Self-supervised training is at best equivalent to supervised training  \cite{lehtinen2018noise2noise,ehret2019model,dewil2021self,nguyen2021self}.
The worse performance of the supervised method can be explained by the residual misalignment of the L1C images (see Sec.~\ref{sec:simulated-ds}) used as ground truth during training, resulting in blurry super-resolved images. The self-supervised method does not suffer from this problem: during training each band of the target image is aligned to the green band of the super-resolved image by the \textbf{CSR} module. It should be possible to compensate the misalignment of the ground truth in a supervised setting by aligning the ground truth bands to the super-resolved prediction (see for example \cite{deudon2002highres}). We did not explore this approach here as our interest lies mainly in the self-supervised training.}

\begin{table}[t!]
    \centering
    \caption{Multi-band super-resolution with the Green band being the reference. PSNRs obtained after aligning GT bands to SR bands. Best PSNR in {\bf bold}.}
    \begin{tabular}{lcccc}
    \toprule
                & \multicolumn{4}{c}{Testing bands}\\ 
                \cmidrule{2-5}
     Training bands & B & G & R & N \tabularnewline
    \midrule    
     G    &                &         46.39  &                &        \tabularnewline
     BG   &         50.87  &         49.77  &                &        \tabularnewline
     BGR  &         51.62  &         51.09  &         48.51  &        \tabularnewline
     BGRN & \textbf{51.80} & \textbf{51.67} & \textbf{48.82} & 41.34  \tabularnewline
    \bottomrule
    \end{tabular}
    \label{tab:SR_ablation_G}
\end{table}

\begin{table}[t!]
    \centering
    \caption{Comparison with supervised training. PSNRs obtained after aligning GT bands to SR bands. Best PSNR in {\bf bold}.}
    \begin{tabular}{lcccc}
    \toprule
                & \multicolumn{4}{c}{Testing bands}\\ 
                \cmidrule{2-5}
     Methods &  B & G & R & N \tabularnewline
     \midrule
     Supervised & 50.90 & 51.04 & 48.14 & 40.98\\
     \SelfSR           & \textbf{51.80} & \textbf{51.67} & \textbf{48.82} & \textbf{41.34}\\
    \bottomrule
    \end{tabular}
    \label{tab:SR_ablation_G_alignment}
\end{table}

\subsection{Real L1B dataset}
\label{sec:l1bdataset}

The Sentinel-2 MSI includes 12 detectors for the VNIR bands arranged on a focal plane. The L1B product is composed of individual rasters, one per detector and per band.
By design, two successive detectors acquire the scene with significant overlap, i.e. 120-200 pixels for the RGBN bands, which allows us to train our model on real L1B data.

However, due to the sensor layout, there is a noticeable vertical offset between consecutive detectors. To prepare the training and testing datasets, we pre-register the bands from different detectors using an integer translation. This process involves estimating a coarse translation between detectors using a SIFT-based matching method, refining the offset for each crop using an optical flow method, and regressing an integer translation.
Since the bands of a given detector are acquired almost simultaneously, it is possible to ensure a registration precision up to a few pixels. For overlaps, registration is less precise due to the time delay, induced parallax and viewpoint changes, but typically remains less than 10 pixels. Overall, {a} refined registration is not required for our framework since the \textbf{CSR} network accurately captures residual shifts.

We used only two Sentinel-2 L1B products in this paper due to their current scarcity, and extracted training data around overlaps, consisting of 3740 pairs of height 400 pixels and width depending on the overlap width between detectors for RGBN bands.\footnote{Note that L1B products will be systematically available through the Copernicus Data Space Ecosystem starting October 2023.}
For testing we extracted crops outside of the overlapping regions, typically near the center of the detectors.

\subsection{Qualitative analysis}
\label{sec:qualitative}

We train the \textbf{CSR} and \textbf{REC} networks sequentially as described in Sec.~\ref{sec:method} on the L1B dataset.
{For \textbf{REC}, the $\ell_{\text{Self-SR}}^*$ loss is used.}
The reconstruction results over the images of the test set are shown in Fig.~\ref{fig:real_self_sr_results}. The \textbf{REC} network is able to successfully restore a high-quality HR image with aligned bands and fine details recovered.

\begin{figure}
    \def\s{0.495\linewidth}
    \centering
    \begin{subfigure}{\s}
    \includegraphics[width=\linewidth]{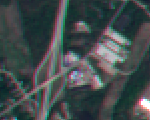}
    \caption{L1B}\label{fig:us-vs-igarss-l1b}
    \end{subfigure}
    \begin{subfigure}{\s}
    \includegraphics[width=\linewidth]{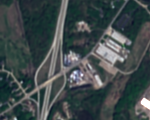}
    \caption{\SelfSR}\label{fig:us-vs-igarss-l1b-sr}
    \end{subfigure}
    \begin{subfigure}{\s}
    \includegraphics[width=\linewidth]{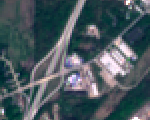}\caption{L1C}\label{fig:us-vs-igarss-l1c}
    \end{subfigure}
    \begin{subfigure}{\s}
    \includegraphics[width=\linewidth]{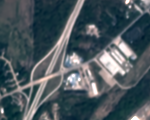}
    \caption{L1C-based SR~\cite{nguyen2023role}}\label{fig:us-vs-igarss-l1c-sr}
    \end{subfigure}
        
        \vspace{-.5em}
        
    \caption{SR (x2) reconstruction by our proposed method~\subref{fig:us-vs-igarss-l1b-sr} using L1B and the method in~\cite{nguyen2023role}~\subref{fig:us-vs-igarss-l1c-sr} using L1C imagery. \subref{fig:us-vs-igarss-l1b} and \subref{fig:us-vs-igarss-l1c} are from the same acquisition.}
    \label{fig:us-vs-igarss}
\end{figure}

To evaluate our self-supervised method, we compare it against another $\times 2$ SISR method~\cite{nguyen2023role}. The method in~\cite{nguyen2023role} relies on ground truth data obtained from PlanetScope imagery. It is trained with a $L_1$ loss, and is designed to work with Sentinel-2 L1C products. For comparison, we identified the L1C products from the same acquisition as the L1B samples and extracted the corresponding crops.
Fig.~\ref{fig:us-vs-igarss} shows the SR results of the proposed methods on L1B, and of~\cite{nguyen2023role} on L1C.
We found that the recovered details are very similar, yet our approach does not suffer from the gap between Sentinel-2 and PlanetScope images, which results in radiometric and geometric degradation by the network.

\section{Conclusion}

To conclude, this paper presents a novel self-supervised method, \SelfSR{}, for single-image super-resolution of Sentinel-2 L1B 10m bands. Leveraging the unique hardware design of Sentinel-2, the proposed method achieves remarkable performance without the need for HR ground truth images. By training a cross-spectral registration module using an innovative self-supervised loss, \DWL{}, \SelfSR{} is capable of reconstructing high-resolution outputs with all bands correctly registered.
To the best of our knowledge, our work is the first to explore the overlapping regions between detectors of Sentinel-2 and their potential for self-supervised SISR.
Through an ablation study on a synthetic dataset and comparison with other supervised SISR methods for Sentinel-2 L1C images, we have demonstrated that \SelfSR{} achieves performance on par with that of supervised training. The availability of our training dataset on the project website makes it a valuable resource for various image restoration and cross-spectral registration tasks.

\paragraph{Acknowledgements}
This work was supported by a grant from Région Île-de-France. It was partly financed by the DGA Astrid Maturation project SURECAVI ANR-21-ASM3-0002. It was  performed using HPC resources from GENCI–IDRIS (grants 2023-AD011012453R2, 2023-AD011012458R2, 2023-AD011012472R2). We thank CNES and ESA for providing the two Sentinel-2 L1B products used in this work.
Centre Borelli is also with Université Paris Cité, SSA and INSERM.

{\small
\bibliographystyle{ieee_fullname}
\bibliography{main.bbl}

\begin{thebibliography}{10}\itemsep=-1pt

\bibitem{anger2020fast}
J{\'e}r{\'e}my Anger, Thibaud Ehret, Carlo de Franchis, and Gabriele Facciolo.
\newblock Fast and accurate multi-frame super-resolution of satellite images.
\newblock {\em ISPRS Annals of Photogrammetry, Remote Sensing \& Spatial
  Information Sciences}, 5(1), 2020.

\bibitem{arar2020unsupervised}
Moab Arar, Yiftach Ginger, Dov Danon, Amit~H Bermano, and Daniel Cohen-Or.
\newblock Unsupervised multi-modal image registration via geometry preserving
  image-to-image translation.
\newblock In {\em Proceedings of the IEEE/CVF conference on computer vision and
  pattern recognition}, pages 13410--13419, 2020.

\bibitem{arefin2020multi}
Md~Rifat Arefin, Vincent Michalski, Pierre-Luc St-Charles, Alfredo Kalaitzis,
  Sookyung Kim, Samira~E. Kahou, and Yoshua Bengio.
\newblock Multi-image super-resolution for remote sensing using deep recurrent
  networks.
\newblock In {\em Proceedings of the IEEE/CVF Conference on Computer Vision and
  Pattern Recognition Workshops}, pages 206--207, 2020.

\bibitem{baker02limits}
Simon Baker and Takeo Kanade.
\newblock Limits on super-resolution and how to break them.
\newblock {\em {IEEE} Transactions on Pattern Analysis and Machine
  Intelligence}, 24(9):1167--1183, 2002.

\bibitem{cai2019toward}
Jianrui Cai, Hui Zeng, Hongwei Yong, Zisheng Cao, and Lei Zhang.
\newblock Toward real-world single image super-resolution: A new benchmark and
  a new model.
\newblock In {\em Proceedings of the IEEE/CVF International Conference on
  Computer Vision}, pages 3086--3095, 2019.

\bibitem{cao2020boosting}
Si-Yuan Cao, Hui-Liang Shen, Shu-Jie Chen, and Chunguang Li.
\newblock Boosting structure consistency for multispectral and multimodal image
  registration.
\newblock {\em IEEE Transactions on Image Processing}, 29:5147--5162, 2020.

\bibitem{deudon2002highres}
Michel Deudon, Alfredo Kalaitzis, Israel Goytom, Md~Rifat Arefin, Zhichao Lin,
  Kris Sankaran, Vincent Michalski, Samira~E Kahou, Julien Cornebise, and
  Yoshua Bengio.
\newblock Highres-net: Recursive fusion for multi-frame super-resolution of
  satellite imagery. arxiv 2020.
\newblock {\em arXiv preprint arXiv:2002.06460}, 2020.

\bibitem{dewil2021self}
Val{\'e}ry Dewil, J{\'e}r{\'e}my Anger, Axel Davy, Thibaud Ehret, Gabriele
  Facciolo, and Pablo Arias.
\newblock Self-supervised training for blind multi-frame video denoising.
\newblock In {\em Proceedings of the IEEE/CVF Winter Conference on Applications
  of Computer Vision}, pages 2724--2734, 2021.

\bibitem{drusch2012sentinel}
Matthias Drusch, Umberto Del~Bello, S{\'e}bastien Carlier, Olivier Colin,
  Veronica Fernandez, Ferran Gascon, Bianca Hoersch, Claudia Isola, Paolo
  Laberinti, Philippe Martimort, et~al.
\newblock Sentinel-2: Esa's optical high-resolution mission for gmes
  operational services.
\newblock {\em Remote sensing of Environment}, 120:25--36, 2012.

\bibitem{ehret2019join}
Thibaud Ehret, Axel Davy, Pablo Arias, and Gabriele Facciolo.
\newblock Joint demosaicing and denoising by overfitting of bursts of raw
  images.
\newblock In {\em The IEEE International Conference on Computer Vision (ICCV)},
  2019.

\bibitem{ehret2019model}
Thibaud Ehret, Axel Davy, Jean-Michel Morel, Gabriele Facciolo, and Pablo
  Arias.
\newblock Model-blind video denoising via frame-to-frame training.
\newblock In {\em The IEEE Conference on Computer Vision and Pattern
  Recognition (CVPR)}, June 2019.

\bibitem{galar2020super}
Mikel Galar, Rub{\'e}n Sesma, Christian Ayala, Lourdes Albizua, and Carlos
  Aranda.
\newblock Super-resolution of sentinel-2 images using convolutional neural
  networks and real ground truth data.
\newblock {\em Remote Sensing}, 12(18):2941, 2020.

\bibitem{gargiulo2019fast}
Massimiliano Gargiulo, Antonio Mazza, Raffaele Gaetano, Giuseppe Ruello, and
  Giuseppe Scarpa.
\newblock Fast super-resolution of 20 m sentinel-2 bands using convolutional
  neural networks.
\newblock {\em Remote Sensing}, 11(22):2635, 2019.

\bibitem{gascon2017copernicus}
Ferran Gascon, Catherine Bouzinac, Olivier Th{\'e}paut, Mathieu Jung, Benjamin
  Francesconi, J{\'e}r{\^o}me Louis, Vincent Lonjou, Bruno Lafrance,
  St{\'e}phane Massera, Ang{\'e}lique Gaudel-Vacaresse, et~al.
\newblock Copernicus sentinel-2a calibration and products validation status.
\newblock {\em Remote Sensing}, 9(6):584, 2017.

\bibitem{glorot2010understanding}
Xavier Glorot and Yoshua Bengio.
\newblock Understanding the difficulty of training deep feedforward neural
  networks.
\newblock In {\em Proceedings of the thirteenth international conference on
  artificial intelligence and statistics}, pages 249--256. JMLR Workshop and
  Conference Proceedings, 2010.

\bibitem{hu2018squeeze}
Jie Hu, Li Shen, and Gang Sun.
\newblock Squeeze-and-excitation networks.
\newblock In {\em Proceedings of the IEEE conference on computer vision and
  pattern recognition}, pages 7132--7141, 2018.

\bibitem{isola2017image}
Phillip Isola, Jun-Yan Zhu, Tinghui Zhou, and Alexei~A Efros.
\newblock Image-to-image translation with conditional adversarial networks.
\newblock In {\em Proceedings of the IEEE conference on computer vision and
  pattern recognition}, pages 1125--1134, 2017.

\bibitem{jiang2021review}
Xingyu Jiang, Jiayi Ma, Guobao Xiao, Zhenfeng Shao, and Xiaojie Guo.
\newblock A review of multimodal image matching: Methods and applications.
\newblock {\em Information Fusion}, 73:22--71, 2021.

\bibitem{kim2020unsupervised}
Gwantae Kim, Jaihyun Park, Kanghyu Lee, Junyeop Lee, Jeongki Min, Bokyeung Lee,
  David~K Han, and Hanseok Ko.
\newblock Unsupervised real-world super resolution with cycle generative
  adversarial network and domain discriminator.
\newblock In {\em Proceedings of the IEEE/CVF Conference on Computer Vision and
  Pattern Recognition Workshops}, pages 456--457, 2020.

\bibitem{kingma2014adam}
Diederik~P Kingma and Jimmy Ba.
\newblock Adam: A method for stochastic optimization.
\newblock {\em arXiv preprint arXiv:1412.6980}, 2014.

\bibitem{lafenetre2023handheld}
Jamy Lafenetre, Ngoc~Long Nguyen, Gabriele Facciolo, and Thomas Eboli.
\newblock Handheld burst super-resolution meets multi-exposure satellite
  imagery.
\newblock {\em arXiv preprint arXiv:2303.05879}, 2023.

\bibitem{lanaras2018super}
Charis Lanaras, Jos{\'e} Bioucas-Dias, Silvano Galliani, Emmanuel Baltsavias,
  and Konrad Schindler.
\newblock Super-resolution of {S}entinel-2 images: Learning a globally
  applicable deep neural network.
\newblock {\em ISPRS Journal of Photogrammetry and Remote Sensing},
  146:305--319, 2018.

\bibitem{lee2009learning}
Daewon Lee, Matthias Hofmann, Florian Steinke, Yasemin Altun, Nathan~D Cahill,
  and Bernhard Scholkopf.
\newblock Learning similarity measure for multi-modal 3d image registration.
\newblock In {\em 2009 IEEE Conference on Computer Vision and Pattern
  Recognition}, pages 186--193. IEEE, 2009.

\bibitem{lehtinen2018noise2noise}
Jaakko Lehtinen, Jacob Munkberg, Jon Hasselgren, Samuli Laine, Tero Karras,
  Miika Aittala, and Timo Aila.
\newblock {Noise2Noise: Learning Image Restoration without Clean Data}.
\newblock In {\em 35th International Conference on Machine Learning, ICML
  2018}, 2018.

\bibitem{liu2018novel}
Xiangzeng Liu, Yunfeng Ai, Juli Zhang, and Zhuping Wang.
\newblock A novel affine and contrast invariant descriptor for infrared and
  visible image registration.
\newblock {\em Remote Sensing}, 10(4):658, 2018.

\bibitem{lugmayr2019unsupervised}
Andreas Lugmayr, Martin Danelljan, and Radu Timofte.
\newblock Unsupervised learning for real-world super-resolution.
\newblock In {\em 2019 IEEE/CVF International Conference on Computer Vision
  Workshop (ICCVW)}, pages 3408--3416. IEEE, 2019.

\bibitem{martens2019super}
Marcus M{\"a}rtens, Dario Izzo, Andrej Krzic, and Dani{\"e}l Cox.
\newblock Super-resolution of proba-v images using convolutional neural
  networks.
\newblock {\em Astrodynamics}, 3:387--402, 2019.

\bibitem{michel2022sen2venmus}
Julien Michel, Juan Vinasco-Salinas, Jordi Inglada, and Olivier Hagolle.
\newblock Sen2ven$\mu$s, a dataset for the training of sentinel-2
  super-resolution algorithms.
\newblock {\em Data}, 7(7):96, 2022.

\bibitem{nguyen2021self}
Ngoc~Long Nguyen, J{\'e}r{\'e}my Anger, Axel Davy, Pablo Arias, and Gabriele
  Facciolo.
\newblock Self-supervised multi-image super-resolution for push-frame satellite
  images.
\newblock In {\em Proceedings of the IEEE/CVF Conference on Computer Vision and
  Pattern Recognition}, pages 1121--1131, 2021.

\bibitem{nguyen2022self}
Ngoc~Long Nguyen, J{\'e}r{\'e}my Anger, Axel Davy, Pablo Arias, and Gabriele
  Facciolo.
\newblock Self-supervised super-resolution for multi-exposure push-frame
  satellites.
\newblock In {\em Proceedings of the IEEE/CVF Conference on Computer Vision and
  Pattern Recognition}, pages 1858--1868, 2022.

\bibitem{nguyen2023role}
Ngoc~Long Nguyen, J{\'e}r{\'e}my Anger, Lara Raad, Bruno Galerne, and Gabriele
  Facciolo.
\newblock On the role of alias and band-shift for sentinel-2 super-resolution.
\newblock {\em arXiv preprint arXiv:2302.11494}, 2023.

\bibitem{perez2013tv}
Javier~S{\'a}nchez P{\'e}rez, Enric Meinhardt-Llopis, and Gabriele Facciolo.
\newblock Tv-l1 optical flow estimation.
\newblock {\em Image Processing On Line}, 2013:137--150, 2013.

\bibitem{pielawski2020comir}
Nicolas Pielawski, Elisabeth Wetzer, Johan {\"O}fverstedt, Jiahao Lu, Carolina
  W{\"a}hlby, Joakim Lindblad, and Natasa Sladoje.
\newblock Comir: Contrastive multimodal image representation for registration.
\newblock {\em Advances in neural information processing systems},
  33:18433--18444, 2020.

\bibitem{pineda2020generative}
Ferdinand Pineda, Victor Ayma, and César Beltran.
\newblock A generative adversarial network approach for super-resolution of
  sentinel-2 satellite images.
\newblock {\em The International Archives of Photogrammetry, Remote Sensing and
  Spatial Information Sciences}, 43:9--14, 2020.

\bibitem{ronneberger2015u}
Olaf Ronneberger, Philipp Fischer, and Thomas Brox.
\newblock U-net: Convolutional networks for biomedical image segmentation.
\newblock In {\em Medical Image Computing and Computer-Assisted
  Intervention--MICCAI 2015: 18th International Conference, Munich, Germany,
  October 5-9, 2015, Proceedings, Part III 18}, pages 234--241. Springer, 2015.

\bibitem{salgueiro2020super}
Luis Salgueiro~Romero, Javier Marcello, and Ver{\'o}nica Vilaplana.
\newblock Super-resolution of {S}entinel-2 imagery using generative adversarial
  networks.
\newblock {\em Remote Sensing}, 12(15):2424, 2020.

\bibitem{sheth2021unsupervised}
Dev~Yashpal Sheth, Sreyas Mohan, Joshua~L. Vincent, Ramon Manzorro, Peter~A.
  Crozier, Mitesh~M. Khapra, Eero~P. Simoncelli, and Carlos Fernandez-Granda.
\newblock Unsupervised deep video denoising.
\newblock In {\em Proceedings of the IEEE/CVF International Conference on
  Computer Vision}, pages 1759--1768, 2021.

\bibitem{shocher2018zero}
Assaf Shocher, Nadav Cohen, and Michal Irani.
\newblock “zero-shot” super-resolution using deep internal learning.
\newblock In {\em Proceedings of the IEEE Conference on Computer Vision and
  Pattern Recognition}, pages 3118--3126, 2018.

\bibitem{soh2020meta}
Jae~Woong Soh, Sunwoo Cho, and Nam~Ik Cho.
\newblock Meta-transfer learning for zero-shot super-resolution.
\newblock In {\em Proceedings of the IEEE/CVF Conference on Computer Vision and
  Pattern Recognition}, pages 3516--3525, 2020.

\bibitem{walters2021there}
Celyn Walters, Oscar Mendez, Mark Johnson, and Richard Bowden.
\newblock There and back again: Self-supervised multispectral correspondence
  estimation.
\newblock In {\em 2021 IEEE International Conference on Robotics and Automation
  (ICRA)}, pages 5147--5154. IEEE, 2021.

\bibitem{wang2021unsupervised}
Wei Wang, Haochen Zhang, Zehuan Yuan, and Changhu Wang.
\newblock Unsupervised real-world super-resolution: A domain adaptation
  perspective.
\newblock In {\em Proceedings of the IEEE/CVF International Conference on
  Computer Vision}, pages 4318--4327, 2021.

\bibitem{wang2020deep}
Zhihao Wang, Jian Chen, and Steven~CH Hoi.
\newblock Deep learning for image super-resolution: A survey.
\newblock {\em IEEE transactions on pattern analysis and machine intelligence},
  43(10):3365--3387, 2020.

\bibitem{xu2022rfnet}
Han Xu, Jiayi Ma, Jiteng Yuan, Zhuliang Le, and Wei Liu.
\newblock Rfnet: Unsupervised network for mutually reinforcing multi-modal
  image registration and fusion.
\newblock In {\em Proceedings of the IEEE/CVF Conference on Computer Vision and
  Pattern Recognition}, pages 19679--19688, 2022.

\bibitem{ye2017robust}
Yuanxin Ye, Jie Shan, Lorenzo Bruzzone, and Li Shen.
\newblock Robust registration of multimodal remote sensing images based on
  structural similarity.
\newblock {\em IEEE Transactions on Geoscience and Remote Sensing},
  55(5):2941--2958, 2017.

\bibitem{yu2020joint}
Songhyun Yu, Bumjun Park, Junwoo Park, and Jechang Jeong.
\newblock Joint learning of blind video denoising and optical flow estimation.
\newblock In {\em Proceedings of the IEEE/CVF Conference on Computer Vision and
  Pattern Recognition Workshops}, pages 500--501, 2020.

\bibitem{yuan2018unsupervised}
Yuan Yuan, Siyuan Liu, Jiawei Zhang, Yongbing Zhang, Chao Dong, and Liang Lin.
\newblock Unsupervised image super-resolution using cycle-in-cycle generative
  adversarial networks.
\newblock In {\em Proceedings of the IEEE Conference on Computer Vision and
  Pattern Recognition Workshops}, pages 701--710, 2018.

\bibitem{zabalza2022super}
Maialen Zabalza and Angela Bernardini.
\newblock Super-resolution of sentinel-2 images using a spectral attention
  mechanism.
\newblock {\em Remote Sensing}, 14(12):2890, 2022.

\bibitem{zhang2018mutual}
Junhao Zhang, Masoumeh Zareapoor, Xiangjian He, Donghao Shen, Deying Feng, and
  Jie Yang.
\newblock Mutual information based multi-modal remote sensing image
  registration using adaptive feature weight.
\newblock {\em Remote Sensing Letters}, 9(7):646--655, 2018.

\bibitem{zhang2018image}
Yulun Zhang, Kunpeng Li, Kai Li, Lichen Wang, Bineng Zhong, and Yun Fu.
\newblock Image super-resolution using very deep residual channel attention
  networks.
\newblock In {\em Proceedings of the European conference on computer vision
  (ECCV)}, pages 286--301, 2018.

\end{thebibliography}
}

\clearpage

\end{document}